\documentclass[10pt,twocolumn,letterpaper]{article}

\usepackage{cvpr}
\usepackage{times}
\usepackage{epsfig}
\usepackage{graphicx}
\usepackage{amsmath}
\usepackage{amssymb}
\usepackage{subfigure}

\usepackage{epsfig}

\usepackage[breaklinks=true,bookmarks=false]{hyperref}

\newcommand{\modelname}[0]{Similarity R-C3D }
\DeclareMathOperator*{\argmax}{arg\,max}
\newcommand{\ssc}{\sigma} % similarity score

\cvprfinalcopy % *** Uncomment this line for the final submission

\def\cvprPaperID{} % *** Enter the CVPR Paper ID here

\begin{document}

%%%%%%%%% TITLE
%\title{Region Convolutional 3D Network for Few-shot Temporal Activity Detection}
\title{Similarity R-C3D for Few-shot Temporal Activity Detection}

%% omit the following lines up until the closing ``}''.
%% Additional authors and addresses can be added with ``\and'',
%% just like the second author.
%% To save space, use either the email address or home page, not both
%\and

\author{Huijuan Xu$^1$ \hspace{4mm} 
Bingyi Kang$^2$   \hspace{4mm}
Ximeng Sun$^3$  \hspace{4mm}
Jiashi Feng$^2$   \hspace{4mm}
Kate Saenko$^3$  \hspace{4mm}
Trevor Darrell$^1$\\
\and
{
%\small
$^1$University of California, Berkeley  \qquad
$^2$National University of Singapore \qquad $^3$Boston University
}\\
\and
{\tt\small 
$^1$\{huijuan, trevor\}@eecs.berkeley.edu, 
%2\{hxu, saenko\}@bu.edu, 
$^3$\{sunxm, saenko\}@bu.edu}
}

\maketitle
%\thispagestyle{empty}

%%%%%%%%% ABSTRACT
\begin{abstract}
Many activities of interest are rare events, with only a few labeled examples available.
Therefore models for temporal activity detection which are able to learn from a few examples are desirable. 
In this paper, we present a conceptually simple and general yet novel framework for few-shot temporal activity detection which detects the start and end time of the few-shot input activities in an untrimmed video.
Our model is end-to-end trainable and can benefit from more few-shot examples.
At test time, each proposal is assigned the label of the few-shot activity class corresponding to the maximum similarity score.
Our Similarity R-C3D method outperforms previous work on three large scale benchmarks for temporal activity detection (THUMOS14, ActivityNet1.2 and ActivityNet1.3 datasets)~in the few-shot setting. Our code will be made available.

\end{abstract}

\section{Introduction}
As the popularity of devices with video recording capabilities rises, the amount of video data tends to explode.
Most of the videos are in untrimmed form and contain few interesting events, while the majority of the content is background. 
Temporal activity detection tries to automatically detect the start and end times of  interesting events in the untrimmed video and enables many real-world applications, e.g. unusual event detection in surveillance video.
The fact that these  events are rare means that models must detect them based on few training examples.

\begin{figure}
    \centering
    \includegraphics[width=\linewidth]{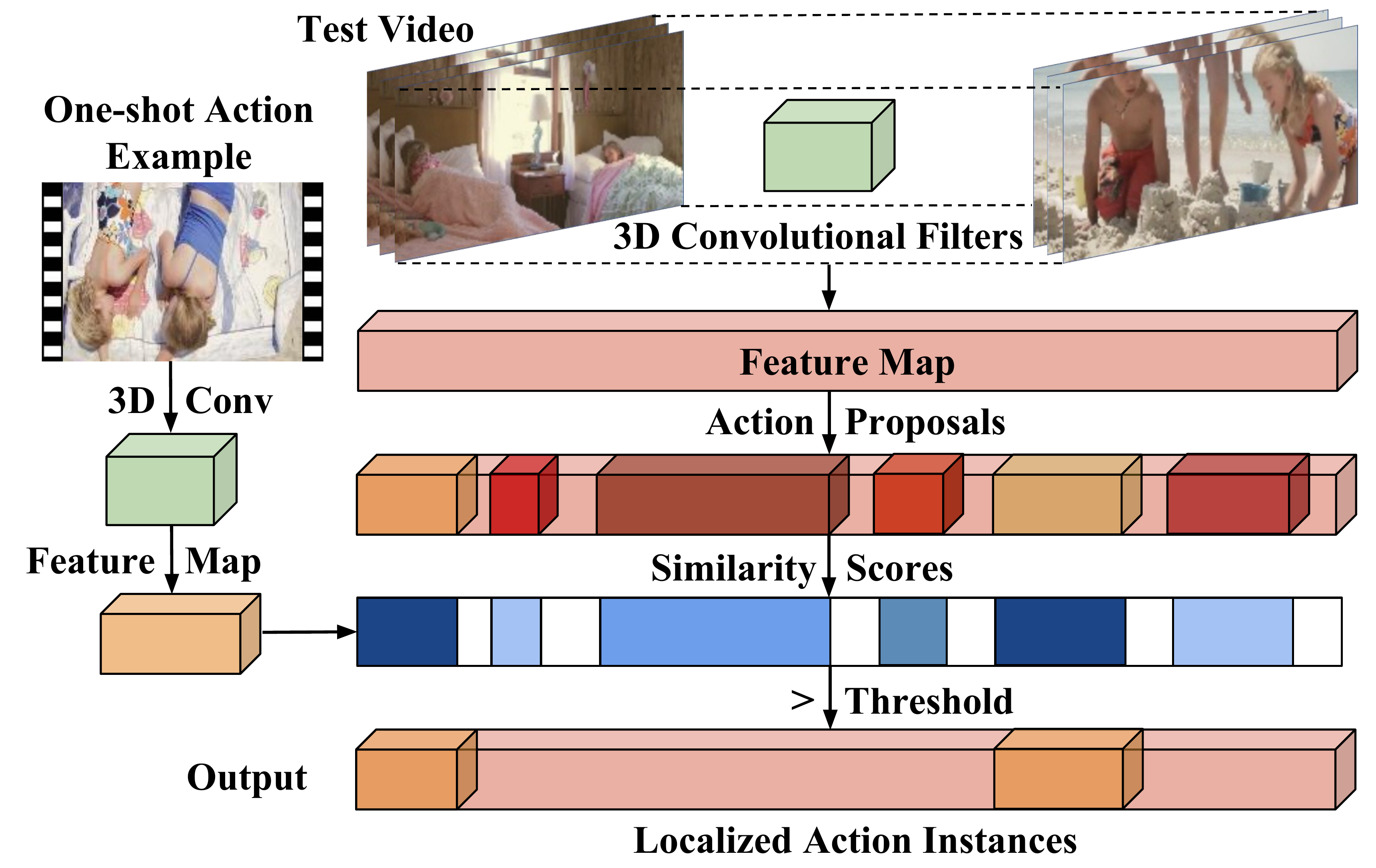}
    \caption{Overview of the proposed \textit{Similarity RC3D} model for few-shot activity detection in untrimmed video. Our model efficiently encodes video with 3D convolutions, proposes foreground segments (action proposals) and assigns labels to them based on their similarity to the one/few-shot training examples of activities.}
    \label{fig:problem_overview}
    \vspace{-10pt}
\end{figure}

Few-shot activity detection~\cite{yang2018one} addresses this problem via example-based action detection with a few training examples as input.
\cite{yang2018one} is based on the Matching Network and utilizes the correlations between the sliding window feature encoding and few-shot example video features to localize actions of previously unseen classes.
However their model uses the ``sliding window'' approach to propose likely activities (proposals), and thus suffers from the high computational cost and inflexible activity boundaries.
To localize proposals for few-shot detection more accurately, in this paper we propose an approach
based on two-stage proposal/refinement networks using direct coordinate regressions for localization, called \textit{Similarity R-C3D}.
Our Similarity R-C3D model (Fig.~\ref{fig:problem_overview}) is derived from the two-stage activity detection framework R-C3D~\cite{xu2017r}, with the second classification stage re-purposed for class-agnostic proposal classification.
The entire untrimmed video is encoded by 3D convolutional (C3D) features~\cite{tran2015learning} on top of which a two-stage proposal network is applied. Refined proposals (foreground events) from the second stage are 
paired with a binary proposal score and then compared with few-shot examples using cosine similarity function.

To extract features for the few-shot example inputs, we design an additional branch which shares weights with the main C3D branch.
After feature extraction, we compare each proposal's features with each few-shot example's features to produce a cosine similarity score, and each proposal is assigned a few-shot class label corresponding to the maximum similarity score.
The final detection is determined by both the proposal score and the similarity score, where the proposal score filters out background activities and the similarity score assigns the few-shot class label.
Our  model is trained following the same meta-learning setting as in~\cite{yang2018one} to enable a fair comparison.

Despite being an end-to-end model, \cite{yang2018one} does not demonstrate the data-driven benefits of end-to-end learning in reported results.
Namely, as the model sees more few-shot example videos, the detection results do not show an obvious improvement.
This unsatisfactory trend might also be related to the localization strategy based on rigid sliding windows adopted by the method~\cite{yang2018one}.
Our model is end-to-end trainable, and demonstrates the advantages of feature learning in an end-to-end fashion.
As more few-shot example videos are observed by the model, the detection results improve significantly.

We perform extensive comparisons of \modelname to previous activity detection methods in the few-shot setting using three publicly available benchmark datasets - THUMOS'14~\cite{THUMOS14}, ActivityNet1.2 and ActivityNet1.3~\cite{caba2015activitynet}, and achieve new state-of-the-art results on all three datasets.

To summarize, the main contributions of our paper are:
\begin{itemize}
    \item Similarity R-C3D, a few-shot activity detection model with a two-stage proposal network for the efficient and accurate detection of potential events;
    \item an end-to-end few-shot branch for feature extraction and similarity calculation from few-shot examples which can benefit from additional examples;
    \item extensive evaluation on three activity detection datasets that demonstrates the general applicability of our model.
\end{itemize}

\begin{figure*}
    \centering
    \includegraphics[width=.97\linewidth]{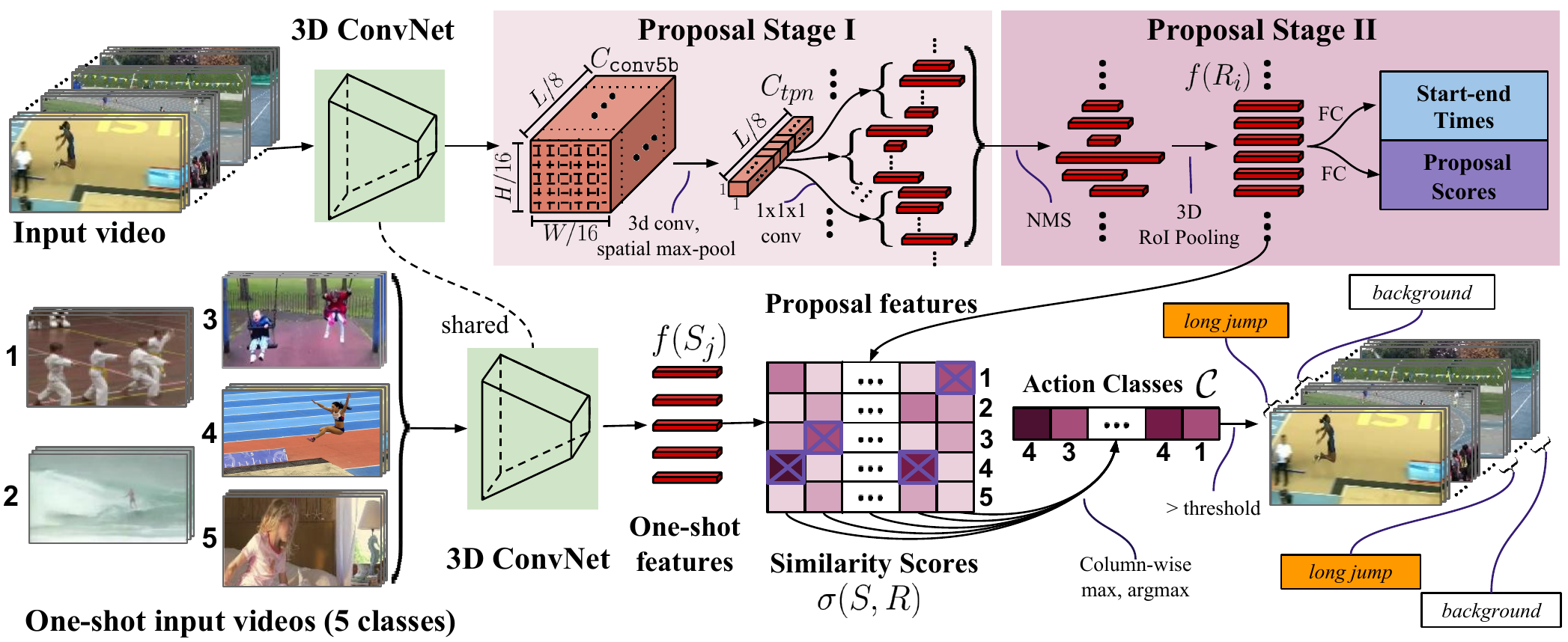}
    \caption{An overview of our proposed Similarity R-C3D network for few-shot activity detection. Given an untrimmed test video, and one- or few-shot training examples of an activity class, the network detects instances of the given activities in the test video and localizes their start and end times. The model has three components: (1) a  3D ConvNet feature extractor applied to both the untrimmed input video and the few-shot trimmed videos, (2) a two-stage temporal proposal network, and (3) a similarity network that computes similarity scores between proposals and few-shot videos and assigns labels.}
    \label{fig:architecture}
    \vspace{-10pt}
\end{figure*}

\section{Related Work}
\label{Related}

%%%%%%%%%%%%%%%%%%%%%%%%%%%%%%%%%%%%%%%%%%%%%%%%%%%%%%%%%%%%%%%%%%%%%%%%
\subsection{Activity Detection}
There are two types of activity detection tasks: spatio-temporal and temporal-only.
Spatio-temporal activity detection localizes activities within spatio-temporal tubes and requires heavier annotation work to collect the training data.
\cite{gkioxari2015finding,saha2016deep,weinzaepfel2015learning,yu2015fast,zhu2017tornado} temporally track bounding boxes corresponding to activities in each frame to realize spatio-temporal activity detection. 
We focus on temporal activity detection~\cite{escorcia2016daps,ma2016learning,montes2016temporal,shou2016temporal,singh2016multi,yeung2016end} which only predicts the start and end times of the activities within long untrimmed videos and classifies the overall activity without spatially localizing people and objects in the frame.

Existing temporal activity detection approaches are dominated by models that use sliding windows to generate segments and subsequently classify them with activity classifiers trained on multiple features~\cite{karaman2014fast, oneata2014lear, shou2016temporal, wang2014action}.
The use of exhaustive sliding windows is computationally inefficient and constrains the boundary of the detected activities to some extent.
Recently, some approaches have bypassed the need for exhaustive sliding window search and proposed to detect activities with arbitrary lengths~\cite{escorcia2016daps, ma2016learning, montes2016temporal, singh2016multi, yeung2016end}. They achieve this by modeling the temporal evolution of activities using RNNs or LSTM networks and predicting an activity label at each time step.
More recently, CDC~\cite{shou2017cdc} and SSN~\cite{zhao2017temporal} propose the bottom-up activity detection by first predicting labels at the frame-/snippet-level and then fusing them.

The R-C3D activity detection pipeline~\cite{xu2017r} encodes the frames with fully-convolutional 3D filters, finds activity proposals, then classifies and refines them based on pooled features within their boundaries.
We adapt the second classification stage to be a proposal refinement stage to obtain more accurate proposals used in our few-shot setting.

% Weakly supervised activity detection
Aside from supervised activity detection,  recent work~\cite{Wang2017UntrimmedNets} has addressed weakly supervised activity localization from data labeled only with video-level class labels by learning attention weights on uniformly sampled proposals.
A similar problem setting is addressed in~\cite{shou2018autoloc}.
Another type of weakly supervised localization of temporal activities is that given an action sequence (without the ground truth temporal annotation for each action) and a paired video, the model is required to localize each action 
within the input video under the action sequential constraint~\cite{richard2016temporal,bojanowski2014weakly,huang2016connectionist}. 
Our paper addresses a different type of partially-supervised task, called few-shot temporal activity detection, where only a few labeled examples are available for each class.

\subsection{Few-shot Learning}
The goal of few-shot learning is to generalize to new examples despite few available training labels. 
The conventional setting tests on novel classes only, while a generalized setting tests on both seen and unseen classes. 
Metric learning methods based on measuring similarity to few-shot example inputs are widely employed by many few-shot learning algorithms that generate good results~\cite{koch2015siamese,shyam2017attentive,snell2017prototypical,vinyals2016matching}.
Similarity values are typically defined on selected attributes or word vectors~\cite{mishra2018generative,zellers2017zero}.
Another line of few-shot learning algorithms augments data by synthesizing examples for the unseen labels~\cite{long2017zero,zhang2018visual}.
Other approaches train deep learning models to learn to update the parameters for few-shot example inputs~\cite{andrychowicz2016learning,ravi2016optimization,bertinetto2016learning}.

Most few-shot models tend to utilize the meta-learning framework for training on a set of datasets with the goal of learning transferable knowledge among datasets~\cite{santoro2016one,mishra2017meta}. MAML~\cite{finn2017model} is one typical example which combines the meta-learner and the learner into one, and directly computes the gradient with respect to the meta-learning objective.
\cite{mettes2017spatial} studies the zero-shot spatial-temporal action localization using spatial-aware object embeddings.
\cite{yang2018one} is the first work proposing the task of temporal activity detection based on few-shot input examples. It performs the activity retrieval for the few-shot example(s) through a sliding window approach and then combines bottom-up retrieval results to obtain the final temporal activity detection.
This work follows the conventional setting of testing on novel examples only, and uses the MAML training strategy.
In our paper, we work along this direction, and propose a few-shot temporal activity detection model based on a two-stage proposal detector to realize more accurate temporal localization.
For a fair comparison, we adopt the same settings as in~\cite{yang2018one}.

\section{Approach}
\label{sec:approach}

In this paper, we tackle the task of few-shot temporal activity detection, which can also be framed as example-based temporal activity detection.
The goal is to locate all instances of 
activities (temporal segments) in an untrimmed test video, given only a few typical examples from a new set of activity classes for training.
We propose a fast and end-to-end model called \textit{Similarity based Region Convolutional 3D Network (Similarity R-C3D)} to solve the few-shot temporal activity detection task. Our model builds on the state-of-the-art R-C3D model for fast activity detection~\cite{xu2017r}, augmenting it with an extra branch for extracting few-shot example video features and a similarity computation subnet.
Our key idea is to assign one of the few-shot class labels to each detected proposal based on the maximum similarity score between each proposal's features and the features of the few-shot example videos.

\subsection{Model Overview}
An overview of our proposed Similarity R-C3D network is illustrated in Figure~\ref{fig:architecture}.
The model consists of three components: a  3D ConvNet feature extractor~\cite{tran2015learning} that encodes both the untrimmed input video and the few-shot trimmed example videos, a two-stage temporal proposal network, and a similarity network for few-shot class classification.
The two-stage proposal subnet predicts temporal segments of variable length that  contain potential activities, while the similarity network classifies these proposals into one of the few-shot activity categories.
The two-stage proposal subnet and the few-shot trimmed video feature extractor share the same C3D weights.

Next, we describe the shared video feature hierarchies in Sec.~\ref{sec:feature}, the two-stage temporal proposal subnet in Sec.~\ref{sec:proposal} and the similarity classification subnet in Sec.~\ref{sec:similarityNetwork}. Sections~\ref{sec:optimization} and \ref{sec:prediction} detail the optimization strategy during training and testing respectively.

\subsection{3D Convolutional Feature Hierarchies}
\label{sec:feature}
We employ the same 3D ConvNet~\cite{tran2015learning} as in the R-C3D model to extract rich spatio-temporal feature hierarchies for our Similarity R-C3D model. 
In our example-based classification setup, the model accepts two types of input videos, namely, the untrimmed video and the trimmed few-shot example videos.
These two types of videos are processed by two network branches that share the same 3D ConvNet weights and extract features for further similarity calculation in Sec.~\ref{sec:similarityNetwork}.

Suppose the input sequence consisting of RGB video frames has  dimensions $\mathbb{R}^{3\times L\times H\times W}$.
The height ($H$) and width ($W$) of the frames are both set to 112 in our experiments, following~\cite{tran2015learning}.
%For the few-shot example videos in 
For the bottom few-shot branch in Figure~\ref{fig:architecture}, we follow the traditional use of the 3D ConvNet~\cite{tran2015learning} to sample frames uniformly from each few-shot example video, and obtain a fixed-dimensional feature vector for each video (see ``One-shot features'' in Figure~\ref{fig:architecture}). The number of frames $L$ for each video is set to $16$ in the few-shot branch.
For the upper branch which encodes the untrimmed input video, we use fully convolutional layers, which means that the number of frames $L$ can be arbitrary. In our experiments we set it to $512$ or $768$, depending on specific datasets.
We adopt the convolutional layers (\texttt{conv1a} to \texttt{conv5b}) of C3D, so a feature map $C_{conv5b}\in \mathbb{R}^{512\times \frac{L}{8} \times \frac{H}{16}\times \frac{W}{16}}$ ($512$ is the channel dimension of the layer \texttt{conv5b}) is produced as the feature encoding of the untrimmed input video.
We use $C_\mathtt{conv5b}$ activations as the shared input to our two-stage proposal subnet in Sec.~\ref{sec:proposal}.

\subsection{Two-stage Proposal Subnet}
\label{sec:proposal}

\textbf{Proposal Stage I:}
The first proposal stage is similar to the proposal stage in the original R-C3D model.
The input to this stage is the feature encoding $C_{conv5b}$ of the untrimmed video. A series of 3D convolutional filters with kernel size $3\!\!\times \!\!3\!\!\times \!\!3$ and a 3D max-pooling filter with kernel size $1\!\times \!\frac{H}{16} \!\!\times \!\!\frac{W}{16}$ are applied to the input to produce a \textit{temporal} only feature map $C_{tpn}\in \mathbb{R}^{512\times \frac{L}{8} \times 1\times 1}$. 
We pre-define a set of anchor segments on this map as fixed multiscale windows centered at $L/8$ and uniformly distributed temporal locations.
The 512-dimensional feature vector at each temporal location in $C_{tpn}$ is used to predict a relative offset $\left \{\delta c_{i},\delta l_{i} \right \}$ to the center location and the length of each anchor segment $\left \{c_{i},l_{i} \right \},\ i\in \left \{1,\cdots,K \right \}$. These anchors, with the predicted center offsets and lengths, define the initial set of temporal proposals.
The network also predicts a binary label (score) that indicates whether each proposal is a foreground activity or background.
Two $1\!\times \!1\!\times \!\!1$ convolutional layers predict the proposal offsets and scores after receiving $C_{tpn}$.

\textbf{Proposal Stage II:}
In the original R-C3D model the second stage classifies the proposals predicted by the first stage into action categories, training on many labeled examples. In our few-shot model we change this second stage into another class-agnostic proposal stage 
which refines the proposal boundaries from stage I.
Here high quality proposals from proposal stage I are selected by score thresholding and Non-Maximum Suppression (NMS).
 We then use 3D RoI pooling~\cite{xu2017r} to extract features with the fixed size for each variable-length proposal from the convolutional features $C_{conv5b}$ (shared with the Proposal Stage I).
The output of the 3D RoI pooling is fed to a series of fully connected layers to extract proposal features.
We then adopt two separate fully-connected layers (a classification and a regression layer) to again classify proposals into fore/background and predict refined start and end times.
For each proposal $i$, the features $f\left( R_i \right)$ extracted from the aforementioned fully-connected layers after the 3D RoI pooling layer are also used in the Similarity Network to compute similarities to the few-shot video features (See Sec.~\ref{sec:similarityNetwork}).

\textbf{Training}:
For training, positive/negative labels are assigned to the anchor segments in Proposal Stage I, and to candidate proposals 
in Proposal Stage II, based on temporal Intersection-over-Union (tIoU) thresholds. All others are held out from training.
We sample balanced batches with a positive/negative ratio of $1\!:\!1$.
For the proposal boundary regression, ground truth activity segments are transformed with respect to either nearby anchor segments or initial proposals from Proposal Stage I.
$t_i^* = \left \{\delta c_i,\delta l_i \right \}$ represents the coordinate transformation of ground truth segments to anchor segments or proposals. The coordinate transformation is computed as follows:
\begin{equation}
%\vspace{-0.1in}
\vspace{-5pt}
\left\{
\begin{gathered}
  \delta c_i = \big(c_i^* - c_i \big) / l_i \\
  \delta l_i = \text{log}\big(l_i^* / l_i \big)
\end{gathered}
\right.
\label{eq:transformation}
% \vspace{-5pt}
\end{equation}
where $c_i$ and $l_i$ are the center location and the length of anchor segments or proposals. Also, we define the same $c_i^*$ and $l_i^*$ for the ground truth activity segments.

\subsection{Similarity Network for Few-shot Classification}
\label{sec:similarityNetwork}
After obtaining proposals from the two-stage proposal network, we use each proposal's features $f\left(R_i\right)$ to compute a similarity score with the few-shot video features $f\left(S_j\right)$. We  produce a Similarity Score matrix $\ssc\left(S,R\right)$ by calculating such a score for each (proposal, few-shot video) pair.
For each proposal $i$ and few-shot video $j$, $\ssc\left(S_j,R_i\right)$ is computed as the cosine similarity between feature embedding vectors $f\left(S_j\right)$ and $f\left(R_i\right)$: %\ie 
$\ssc(S_j,R_i)=\frac{ \big \langle f \big(S_j \big),f \big(R_i \big) \big \rangle}{\|f \big (S_j \big)\|\|f \big(R_i \big)\|}$.

Once we have the Similarity Score matrix $\ssc\left(S,R\right)$ between the proposal features and the few-shot video features, we can label each proposal $i$ with a few-shot class label $\mathcal{C}_i$.
In the one-shot setting, $\argmax$ is directly applied on each column of the Similarity Score matrix $\ssc(S,R)$: $\mathcal{C}_i=\argmax_j\ssc \big(S_j,R_i \big)$.
In the few-shot setting, we first average the similarity scores of the multiple examples belonging to the same class, and then assign $\mathcal{C}_i$ by applying $\argmax$ among few-shot class labels.

\textbf{Training}:
For training the multi-class classifier in the Similarity Network, we follow the same rules as in the Proposal Stage II to assign positive/negative labels, with the only difference being that now the positive label is not a binary %positive 
proposal label but one of the few-shot class labels.

\subsection{Optimization}
\label{sec:optimization}
We train the network by optimizing the binary proposal classification and regression tasks jointly for the two-stage proposal subnet, as well as optimizing for the multi-class classification task in the few-shot Similarity Network.
The softmax loss function $\mathcal{L}_{cls}$ is used for proposal binary classification, and smooth L1 loss function $\mathcal{L}_{reg}$ \cite{girshick2015fast} is used for regression.
Specifically, the objective function $\mathcal{L}_{p1}$ used in the Proposal Stage I is given by:
\small
\begin{equation}
   % \hspace{-2mm} 
    \mathcal{L}_{p1} = \frac{1}{N_{cls}} \sum\limits_i \mathcal{L}_{cls} \big(a_i, a_i^*\big) + \lambda \frac{1}{N_{reg}} \sum\limits_i a_i^* \mathcal{L}_{reg} \big(t_i, t_i^* \big)
%\vspace{-2mm}
%\label{eq:loss}
\end{equation}
\normalsize
where $N_{cls}$ and $N_{reg}$ stand for batch size and the number of anchor/proposal segments, $\lambda$ is the loss trade-off parameter and is set to a value of $1$. For each anchor/proposal segment $i$ in a batch, $a_i$ is its predicted probability, $a_i^*$ is the ground truth, $t_i = \left \{\delta \widehat{c}_i,\delta \widehat{l}_i \right \}$ represents the  relative offset predicted for anchor segments or proposals, and
$t_i^* = \left \{\delta c_i, \delta l_i \right \}$ represents the coordinate transformation of ground truth segments to anchor segments or proposals. 
A similar objective function $\mathcal{L}_{p2}$ is used in Proposal Stage II. 

The softmax loss function $\mathcal{L}_{fewShot\_cls}$ is used in few-shot multi-class classification.
All five losses from Proposal Stage I, Proposal Stage II and few-shot Similarity Network are optimized jointly, resulting in the following total loss:
%$L_{total}$:
\small
\begin{equation}
    %\hspace{-2mm} 
    \mathcal{L}_{total} = \mathcal{L}_{p1} + \mathcal{L}_{p2} + \mathcal{L}_{fewShot\_cls}
%\vspace{-2mm}
%\label{eq:overal_loss}
\end{equation}
\normalsize

\subsection{Prediction}
\label{sec:prediction}
Few-shot temporal activity detection in Similarity R-C3D consists of two steps.

First, the two-stage proposal subnet generates candidate proposals and predicts the start-end time offsets as well as a proposal score for each.
The proposals from Stage I are refined via NMS with a threshold value of 0.7.
After NMS, the selected proposals are fed to the Proposal Stage II network to again perform binary proposal classification, and the boundaries of the predicted proposals are further refined by the regression layer.
The boundary prediction in both proposal stages is in the form of relative displacement of center point and length of segments.
In order to find the start time and end time of the predicted proposals, inverse coordinate transformation to Equation~\ref{eq:transformation} is performed.

Second, proposals from the Proposal Stage II are further used to calculate their similarity scores with the few-shot trimmed video features, and the maximum similarity score and corresponding class label are assigned for each proposal.
In the multiple shot cases, the scores are first averaged within each class, and then max is taken among classes.
Each proposal from Stage II is paired with a proposal score and a few-shot class with maximum similarity score, and the final activity detection is determined by thresholding both scores.

\section{Experiments}
We evaluate \modelname on three large-scale activity detection datasets - THUMOS'14~\cite{THUMOS14}, ActivityNet1.2 and ActivityNet1.3~\cite{caba2015activitynet}.
Sections~\ref{exp:thumos14},
~\ref{exp:activitynet1.3},~\ref{exp:activitynet1.2} 
provide the experimental details and evaluation results on these three datasets.

We use the same few-shot problem setup as in the previous work~\cite{yang2018one}, with the testing classes (novel classes) having no overlap with the training classes.
We split the  classes in each dataset into two subsets, train on one subset  and test on the other subset. The specific split for each dataset is introduced in the corresponding section.
We also use the same meta-learning setup as in \cite{yang2018one} for fair comparison with its results.
During the meta-training stage, in each training iteration, we randomly sample five classes from the subset of training classes, and then for each class we randomly sample one trimmed example video (in the one-shot setting) to constitute a training batch as the input for the few-shot branch. In the five-shot setting, the number of randomly sampled example videos is five for each class, and the final similarity value for each class is averaged by the five example videos of same class.
Notably, the randomly sampled five classes for the few-shot branch should have at least one class overlap with the ground truth activity classes in the randomly sampled untrimmed input video.
The five  sampled activity classes are assigned labels from 0-4 randomly. 

During the meta-testing stage, the test data preparation is the same as in the meta-training stage, but on the subset of testing classes.
Results are shown in terms of mean Average Precision - mAP@$\alpha$ where $\alpha$ denotes Intersection over Union (IoU) threshold, and average mAP at 10 evenly distributed IoU thresholds between 0.5 and 0.95, as is the common practice in the literature.
However in the meta-testing setting, mAP@0.5 and average mAP are calculated in each iteration for one untrimmed test video, and the final results are reported by averaging over 1000 iterations.

%%%%%%%%%%%%%%%%%%%%%%%%%%%%%%%%%%%%%%%%%%%%%%%%%%%%%%%%%%%%%%%%%%%%%%%%%%%%%%%%

%%%%%%%%%%%%%%%%%%%%%%%%%%%%%%%%%%%%%%%%%%%%%%%%%%%%%%%%%%%%%%%%%%%%%%%%%%

\subsection{Experiments on THUMOS'14}
\label{exp:thumos14}
\noindent{\textbf{Experimental Setup}:}
The THUMOS'14 activity detection dataset contains videos of 20 different sport activities.
The training set contains 2765 trimmed videos from UCF101 dataset~\cite{soomro2012ucf101} while the validation and the test sets contain 200 and 213 untrimmed videos respectively.
The 20 sport activities in THUMOS'14 are a subset of the 101 classes in UCF101.
The 20 classes are split into 6 classes (Thumos-val-6) for training and 14 classes (Thumos-test-14) for testing as in~\cite{yang2018one}.
In the THUMOS'14 dataset, the trimmed videos in the few-shot branch come from UCF101.

For the untrimmed input video, a buffer of 512 frames at 25 frames per second (fps) act as inputs to the two-stage proposal network.
We first pretrain the two-stage proposal network using the proposal annotations from Thumos-val-6, starting from the C3D weights trained on Sports-1M released by the authors in~\cite{tran2015learning}.
Then we use the pre-trained two-stage proposal network weights to initialize the entire Similarity R-C3D network, and allow all the layers to be trained with a fixed learning rate of $10^{-4}$.
The proposal score threshold is set as 0.05 and the similarity score threshold as 0.02.

\noindent{\textbf{Results}:}
In Table~\ref{tab:res_thumos14}, we present a comparative evaluation of the one-shot and five-shot temporal activity detection performance of \modelname with existing state-of-the-art approaches~\footnote{For CDC, we use the values reported in \cite{yang2018one}} in terms of mAP@0.5.
From the table, we see that our \modelname model outperforms previous methods by a large margin in both the one-shot setting and the five-shot setting.
The one-shot mAP@0.5 improves over the sliding window based approach proposed by~\cite{yang2018one} by 11.2\%, and the five-shot mAP@0.5 improves by 14.1\% (one-hundred percent relative improvement).
 
\begin{figure*}[t]
    \centering
    \includegraphics[width=.75\linewidth]{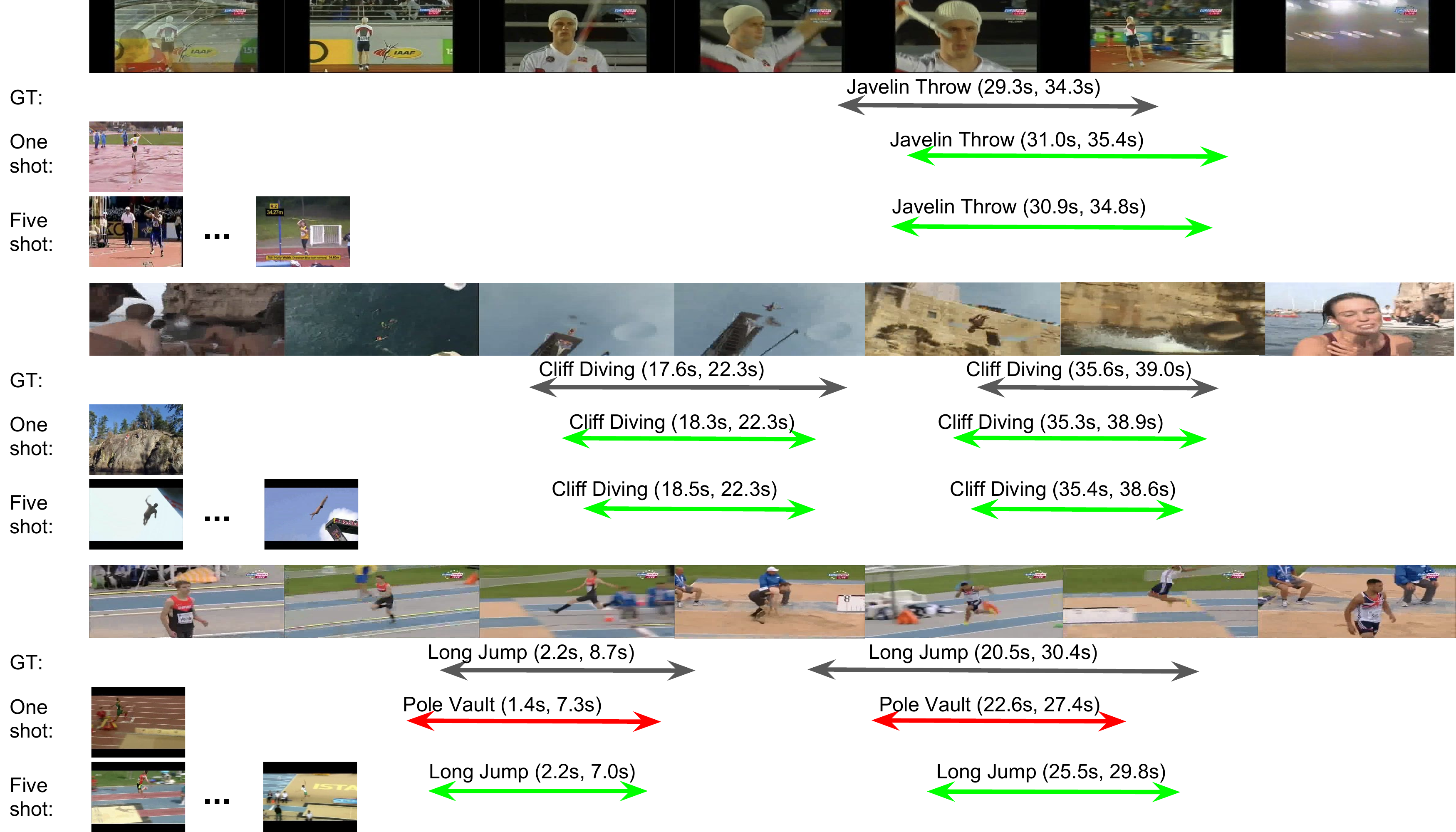}
    \caption{Qualitative visualization of one-shot/five-shot temporal activity detection results from our model, \modelname, on THUMOS'14 dataset. Ground truth (GT) clips corresponding to GT class are marked with black arrows. Correct predictions (predicted clips having tIoU more than 0.5 with ground truth) are marked in green, and incorrect predictions are marked in red. The start and end times are shown in seconds. Best viewed in color.}
    \label{fig:thumos14_examples}
    \vspace{-10pt}
\end{figure*}

Another important finding is that, for the previous methods, the mAP@0.5 result differences between one-shot and five-shot settings are very small, showing almost no improvement from increasing the number of few-shot examples.
However, our method shows a significant improvement from increasing the few-shot examples for each class from one to five, which comes from the data-driven benefit of end-to-end learning in our model.
Figure~\ref{fig:thumos14_examples} shows some representative qualitative results from three videos in this dataset. We see that, compared to one-shot, the five-shot model detects more accurate activity boundaries, and successfully detects the ``Long Jump" activity in the last example while the one-shot model fails and detects the wrong activity class ``Pole Vault".

\begin{table}[!t]
\centering
\caption{Few-shot temporal activity detection results on THUMOS'14 (in percentage). mAP at tIoU threshold $\alpha=0.5$ are reported. @1 means ``one-shot" and @5 means ``five-shot".}
\small
\begin{tabular}{l || c } 
\hline
 ~ & mAP@0.5 \\ \hline
 CDC@1~\cite{shou2017cdc} &  6.4 \\ %\hline
CDC@5~\cite{shou2017cdc} &  6.5 \\ %\hline
 sliding window@1~\cite{yang2018one} &  13.6 \\ %\hline
 sliding window@5~\cite{yang2018one} &  14.0 \\ \hline
  Similarity R-C3D@1  &  24.8  \\ %\hline
  Similarity R-C3D@5  &  28.1  \\ \hline
  \end{tabular}
\label{tab:res_thumos14}
\vspace{-10pt}
\end{table}

%%%%%%%%%%%%%%%%%%%%%%%%%%%%%%%%%%%%%%%%%%%%%%%%%%%%%%%%%%%%%%%%%%%%%%%%
%%%%%%%%%%%%%%%%%%%%%%%%%%%%%%%%%%%%%%%%%%%%%%%%%%%%%%%%%%%%%%%%%%%%%%%%

\begin{figure}
    \centering
    \includegraphics[width=\linewidth]{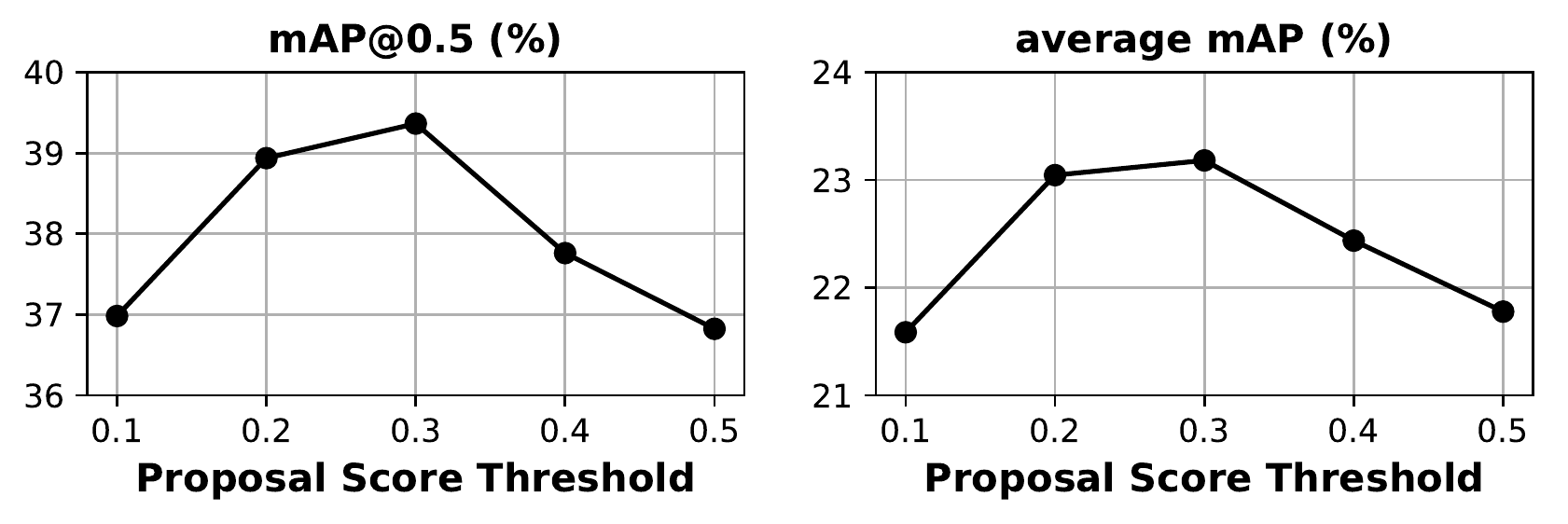}
    \caption{The change of mAP@0.5 and average mAP (in percentage) with the proposal score threshold for the Similarity R-C3D model in one-shot setting on ActivityNet1.2 dataset.}
    \label{fig:mAP_vs_thres}
    \vspace{-20pt}
\end{figure}

%%%%%%%%%%%%%%%%%%%%%%%%%%%%%%%%%%%%%%%%%%%%%%%%%%%%%%%%%%%%%%%%%%%%%%%%%%%%
\subsection{Experiments on ActivityNet1.3}
\label{exp:activitynet1.3}
\noindent{\textbf{Experimental Setup}:}
%%%%%%%%%%%%%%%%%%%%%%%%%%%%%%%%%%%%%%%%%%%%%%%%%%%%%%%%%%%%%%%%%%%%%%%%%%%%%%%%%%%%
The ActivityNet~\cite{caba2015activitynet} dataset consists of untrimmed videos and is released in three versions.
The latest release (1.3) contains 200 different types of activities and around 20K videos. 
Most videos contain activity instances of a single class covering a great deal of the video.
The 200 activity classes are randomly split into 180 classes (ActivityNet1.3-train-180) for training and 20 classes (ActivityNet1.3-test-20) for testing.
In the ActivityNet1.3 dataset, the trimmed videos in the few-shot branch come from the ground truth annotations inside each untrimmed video.

For the two-stage proposal network, the length of the input buffer is set to 768 at 6 fps.
We also pre-train the two-stage proposal network using the proposal annotations from ActivityNet1.3-train-180, starting from the Sports-1M C3D weights.
We initialize the Similarity R-C3D network with the pre-trained two-stage proposal network weights.
As a speed-efficiency trade-off, we freeze the first two convolutional layers in our model during training and the learning rate is fixed at $10^{-4}$.
The proposal score threshold is set as 0.5 and the similarity score threshold as 0.02.

\noindent{\textbf{Results}:}
In Table~\ref{tab:res_act200} we show the performance of \modelname in terms of mAP@0.5 and average mAP. Since no paper has reported results on this dataset before, we are the first to report results in a few-shot temporal activity detection setting on this dataset.
Five-shot outperforms one-shot in both metrics significantly, showing the benefit of our end-to-end model where more few-shot example data is available for training.
Figure~\ref{fig:activitynet1_3_examples} shows some representative qualitative results from this dataset.

\begin{table}[!t]
\centering
\caption{
Few-shot temporal activity detection results on activityNet1.3 (in percentage). mAP at tIoU threshold $\alpha=0.5$ and average mAP of $\alpha \in \{0.5,0.95\}$ are reported. @1 means ``one-shot" and @5 means ``five-shot". }
\small
\begin{tabular}{l || c c} 
\hline
 ~ & mAP@0.5 & average mAP \\ \hline
  Similarity R-C3D@1  &  43.4 & 29.1 \\ %\hline
  Similarity R-C3D@5  &  51.6 & 34.6 \\ \hline
  \end{tabular}
\label{tab:res_act200}
\vspace{-10pt}
\end{table}

%%%%%%%%%%%%%%%%%%%%%%%%%%%%%%%%%%%%%%%%%%%%%%%%%%%%%%%%%%%%%%%%%%%%%%%%%%%%
\subsection{Experiments on ActivityNet1.2}
\label{exp:activitynet1.2}
\noindent{\textbf{Experimental Setup}:}
We also compare our Similarity R-C3D model on the ActivityNet1.2 dataset, as previous few-shot results are reported on this dataset~\cite{yang2018one}.
The ActivityNet1.2 dataset contains around 10k videos and 100 activity classes which are a subset of the 200 activity classes in ActivityNet1.3.
The ActivityNet1.2 dataset and the ActivityNet1.3 dataset share the same video data but have different sets of annotations.
The 100 activity classes are split into 80 classes (ActivityNet1.2-train-80) for training and 20 classes (ActivityNet1.2-test-20) for testing.\footnote{We keep ActivityNet1.2-test-20 and ActivityNet1.3-test-20 the same.}
In the ActivityNet1.2 dataset, the trimmed videos in the few-shot branch also come from the ground truth annotations inside each untrimmed video.
Other settings (the buffer length, proposal weight, pretraining, learning rate, etc.) are set the same as in ActivityNet1.3 dataset (see Sec.~\ref{exp:activitynet1.3}).
The proposal score threshold is set as 0.3 and the similarity score threshold as 0.02.

\begin{figure*}[t]
\centering
\subfigure[ActivityNet1.2 detection examples]{
    \label{fig:activitynet1_2_examples}
    \includegraphics[width=.65\linewidth]{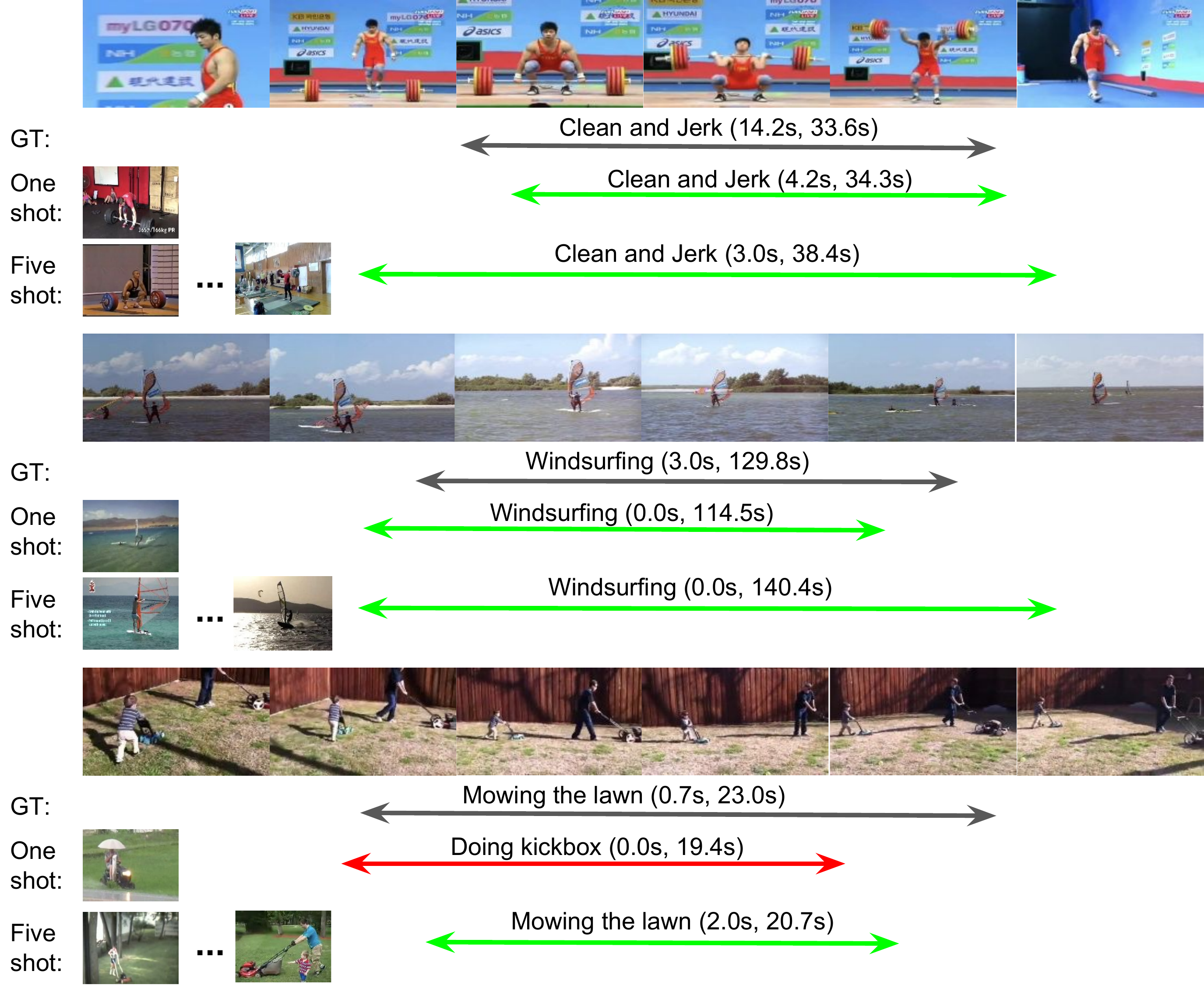}}
\subfigure[ActivityNet1.3 detection examples]{
    \label{fig:activitynet1_3_examples}
    \includegraphics[width=.65\linewidth]{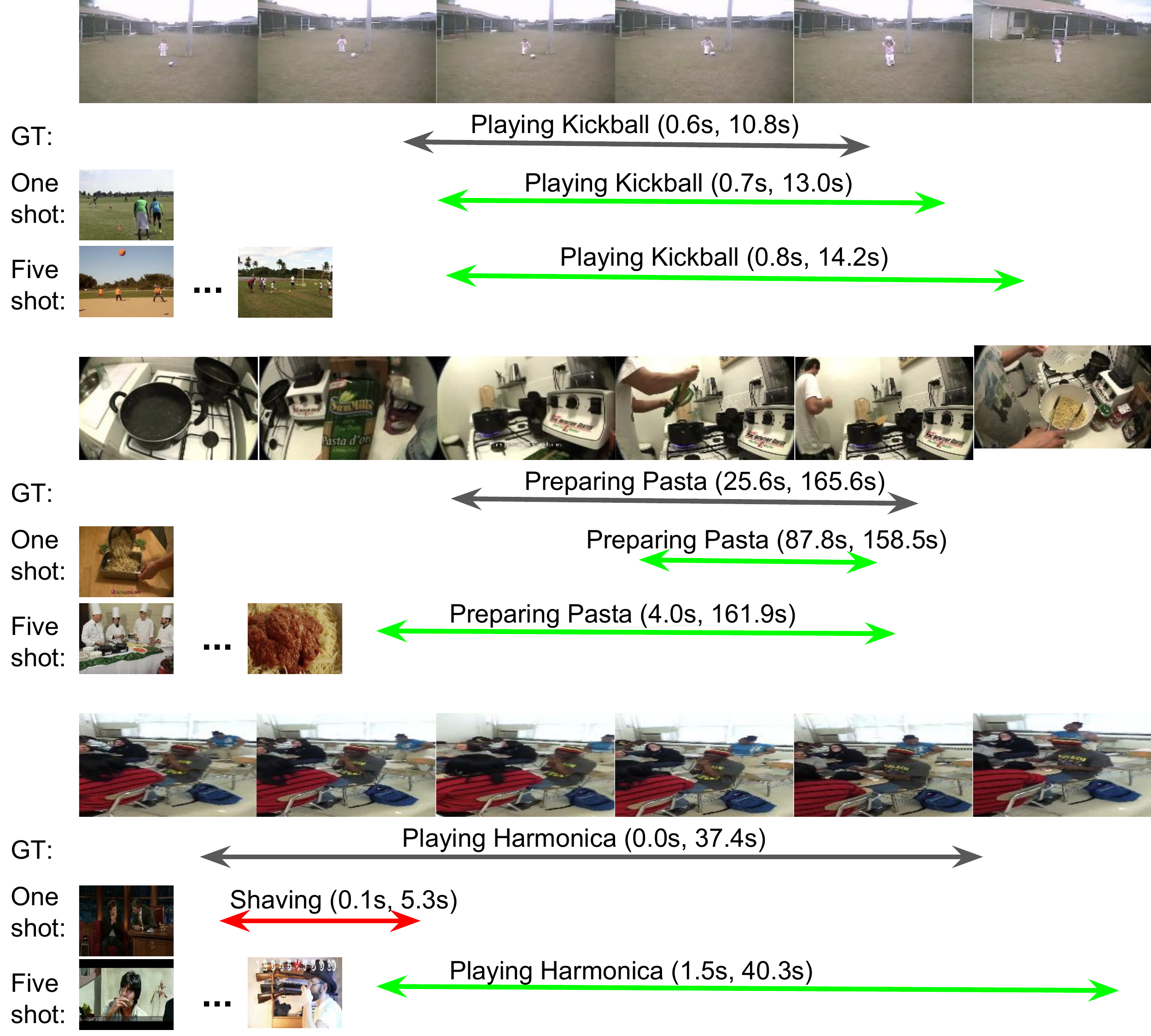}}
\caption
{Qualitative visualization of one-shot/five-shot temporal activity detection results from our model, \modelname, on ActivityNet 1.2 dataset~\subref{fig:activitynet1_2_examples} and ActivityNet 1.3 dataset~\subref{fig:activitynet1_3_examples}. 
Ground truth (GT) clips corresponding to GT class are marked with black arrows. Correct predictions (predicted clips having tIoU more than 0.5 with ground truth) are marked in green, and incorrect predictions are marked in red. The start and end times are shown in seconds. Best viewed in color.}
\label{fig:qualitative}
\end{figure*}

\noindent{\textbf{Results}:}
In Table~\ref{tab:res_act100} we show the performance of \modelname and compare with existing published approaches.
From the results, we see that the CDC model almost fails on this dataset with average mAP around 2\%.
The sliding window based model in ~\cite{yang2018one} gets significantly better results in both mAP@0.5 and average mAP compared to the CDC model.
Our Similarity R-C3D in both one-shot (Similarity R-C3D@1) and five-shot (Similarity R-C3D@5) settings further improves the results by a large margin.
We also observe a significant improvement from increasing the few-shot input from one to five, while the previous two methods (CDC and sliding window based method) do not exhibit such a trend.
Figure~\ref{fig:activitynet1_2_examples} shows some representative qualitative results from this dataset.

\noindent{\textbf{Ablation Study for Proposal Network Pretrain}:}
To investigate the effect of the proposal pretraining, we train another set of one-shot and five-shot experiments on ActivityNet1.2 dataset, starting from the two-stage proposal weights pre-trained on ActivityNet1.3 dataset (namely, ActivityNet1.3-train-180).
The results are shown in Table~\ref{tab:res_act100} with $^*$ mark.
Comparing the corresponding pair of one-shot/five-shot experiments, we can see that the two-stage proposal weights have minor effects on the detection results.
Our Similarity R-C3D model mainly learns the knowledge from the few-shot data during the end-to-end training.

\noindent{\textbf{Ablation Study for Proposal Score Threshold}:}
Recall that our final few-shot detection results are determined by thresholding over the proposal score and similarity score.
To show the effect of the proposal score threshold, we fix the similarity score at 0.02, and plot mAP@0.5 and average mAP with respect to the proposal score threshold for our one-shot setting on the ActivityNet1.2 dataset in Figure~\ref{fig:mAP_vs_thres}.
In this setting, the optimal proposal score threshold is 0.3. This also shows the trade-off 
between the quantity and the quality of proposals.

%%%%%%%%%%%%%%%%%%%%%%%%%%%%%%%%%%%%%%%%%%%%%%%%%%%%%%%%%

\begin{table}[!t]
\centering
\caption{
%Activity detection results on activityNet1.2
Few-shot temporal activity detection results on ActivityNet1.2 (in percentage). mAP at tIoU threshold $\alpha=0.5$ and average mAP of $\alpha \in \{0.5,0.95\}$ are reported. @1 means ``one-shot" and @5 means ``five-shot". The results with $^*$ use the activityNet1.3 pretrained two-stage proposal weights as initialization.
}
\small
\begin{tabular}{l || c c} 
\hline
 ~ & mAP@0.5 & average mAP \\ \hline
 CDC@1~\cite{shou2017cdc} &  8.2 & 2.4\\ %\hline
CDC@5~\cite{shou2017cdc} &  8.6 & 2.5 \\ %\hline
 sliding window@1~\cite{yang2018one} &  22.3 & 9.8 \\ %\hline
  sliding window@5~\cite{yang2018one} &  23.1 & 10.0  \\ \hline
  Similarity R-C3D@1  &  39.4  & 23.2  \\ %\hline
  Similarity R-C3D@5  &  45.8 & 28.2 \\ \hline
  {Similarity R-C3D@1}$^*$  &  39.2  & 24.5  \\ %\hline
  {Similarity R-C3D@5}$^*$  &  45.8 & 27.9 \\ \hline
  \end{tabular}
\label{tab:res_act100}
\vspace{-15pt}
\end{table}

%%%%%%%%%%%%%%%%%%%%%%%%%%%%%%%%%%%%%%%%%%%%%%%%%%%%%%%%%

\section{Conclusion}

In this paper, we propose a few-shot temporal activity detection model, called Similarity R-C3D, which is composed of a two-stage proposal subnet and a few-shot classification branch based on similarity values.
Our model is end-to-end trainable and can benefit from observing additional few-shot examples.
The approach is simple, yet outperforms previous models by a large margin.
One possible future direction is to consider the domain differences between the few-shot examples  and the untrimmed input video, since in the THUMOS14 dataset the few-shot examples come from the UCF101 dataset which is a different visual domain.

% \newpage
\clearpage

{\small
\bibliographystyle{ieee}
\bibliography{egbib}
}

\end{document}